\documentclass[letterpaper]{article} 
\usepackage[submission]{aaai25}  
\usepackage{times}  
\usepackage{helvet}  
\usepackage{courier}  
\usepackage[hyphens]{url}  
\usepackage{graphicx} 
\usepackage{hyperref}
\usepackage{booktabs}
\usepackage{cleveref}
\usepackage{graphicx}  
\usepackage{float}  
\usepackage{subfigure}  
\usepackage{subcaption}
\usepackage{multirow}
\usepackage{xcolor}
\usepackage{colortbl}  
\usepackage{array}   
\definecolor{deepgreen}{rgb}{0.0, 0.5, 0.0}
\usepackage{natbib}  
\usepackage{caption} 
\usepackage{algorithm}
\usepackage{algorithmic}

\usepackage{newfloat}
\usepackage{listings}
\DeclareCaptionStyle{ruled}{labelfont=normalfont,labelsep=colon,strut=off} 
\floatstyle{ruled}
\newfloat{listing}{tb}{lst}{}
\floatname{listing}{Listing}
\title{Image Regeneration: Evaluating Text-to-Image Model via Generating Identical Image with Multimodal Large Language Models}
\author{
    Chutian Meng\textsuperscript{\rm 1}
    Fan Ma\textsuperscript{\rm 1}
    Jiaxu Miao\textsuperscript{\rm 1}
    Chi Zhang\textsuperscript{\rm 1}
    Yi Yang\textsuperscript{\rm 1}
    Yueting Zhuang\textsuperscript{\rm 1}\thanks{Corresponding author.}\\
}
\affiliations{
    \textsuperscript{\rm 1}College of Computer Science and Technology, Zhejiang University\\

}

\usepackage{bibentry}

\begin{document}

\maketitle

\begin{abstract}
Diffusion models have revitalized the image generation domain, playing crucial roles in both academic research and artistic expression. With the emergence of new diffusion models, assessing the performance of text-to-image models has become increasingly important.
Current metrics focus on directly matching the input text with the generated image, but due to cross-modal information asymmetry, this leads to unreliable or incomplete assessment results. Motivated by this, we introduce the Image Regeneration task in this study to assess text-to-image models by tasking the T2I model with generating an image according to the reference image.
We use GPT4V to bridge the gap between the reference image and the text input for the T2I model, allowing T2I models to understand image content.
This evaluation process is simplified as comparisons between the generated image and the reference image are straightforward. Two regeneration datasets spanning content-diverse and style-diverse evaluation dataset are introduced to evaluate the leading diffusion models currently available.
Additionally, we present ImageRepainter framework to enhance the quality of generated images by improving content comprehension via MLLM guided iterative generation and revision.
Our comprehensive experiments have showcased the effectiveness of this framework in assessing the generative capabilities of models. By leveraging MLLM, we have demonstrated that a robust T2M can produce images more closely resembling the reference image.
\end{abstract}

%
\begin{figure}[t]
  \centering
  \includegraphics[width=\linewidth]{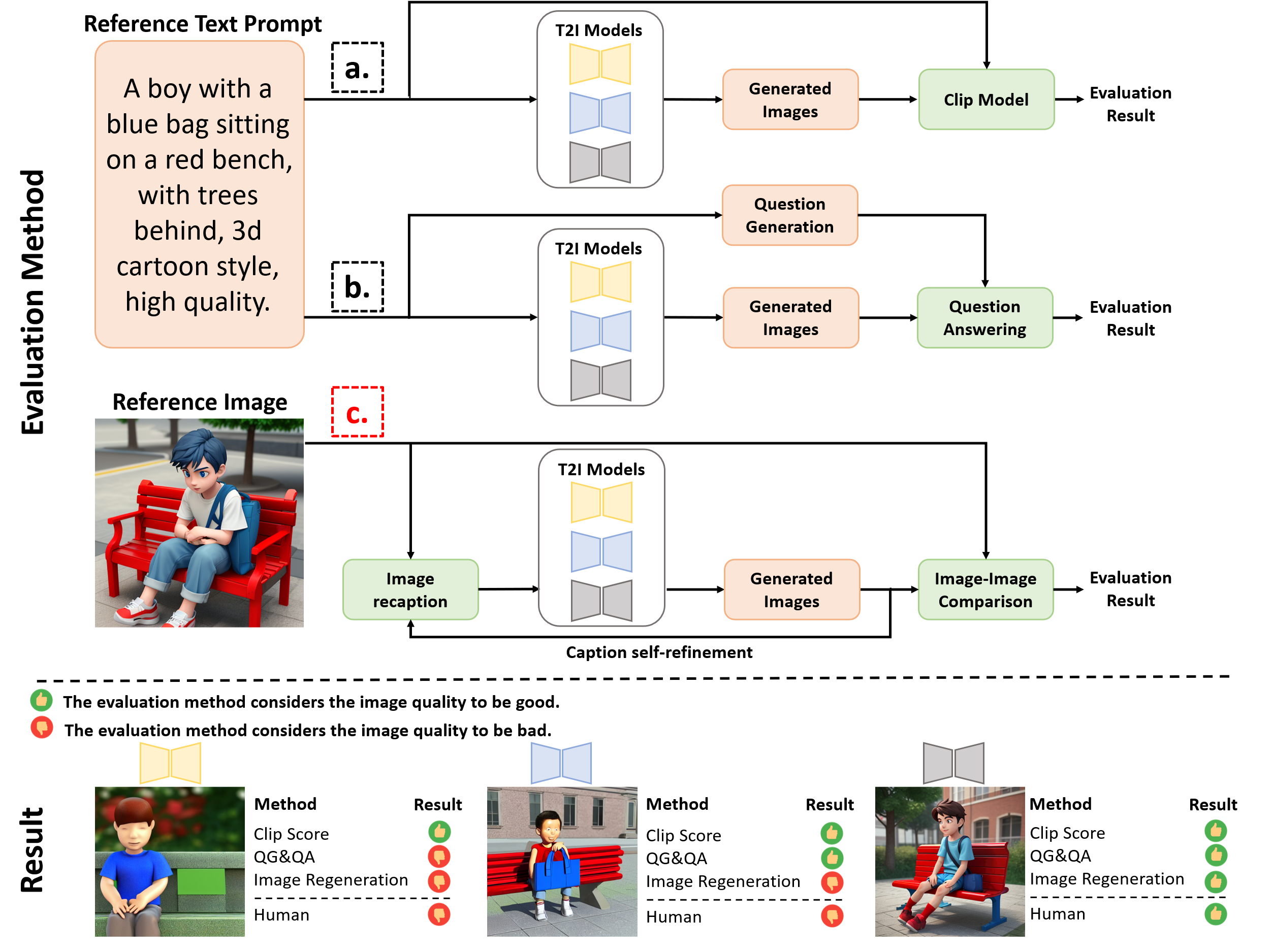}
  \caption{Architecture comparison among (a) pre-trained model evaluation, (b) QG\&QA(Question Generation \& Question Answering) evaluation, (c) our image regeneration evaluation, where our approach (c) achieves better alignment with human cognition. }
  \label{fig:comparch}
\end{figure}

\section{Introduction}
With the rise of generative AI, there has been a surge in the development of text-to-image \cite{rombach2022highresolution,podell2023sdxl,ramesh2022hierarchical,saharia2022photorealistic,yu2022scaling} and text-to-video \cite{khachatryan2023text2videozero,liu2024sora} models in recent years. The primary goal of such algorithms is to generate visual content based on user prompts. These technologies have broad applications \cite{yang2024diffusion}, such as style transfer \cite{zhang2023inversionbased,wang2023stylediffusion}, image editing \cite{gal2022image,ruiz2023dreambooth,ruiz2023hyperdreambooth}, and more, holding significant importance in academic and creative domains.

Despite rapid advancements in related algorithms and applications, a research gap remains in evaluating the quality of these generative models. Current T2I model evaluation focuses on two modalities: text input and image output. For example, CLIP score \cite{radford2021learning} is widely used to measure the semantic information within images based on user prompts. However, the CLIP score has lower sensitivity to fine-grained details and visual variances and cannot reflect the quality of generated images. Similarly, the recently popular evaluation method QG\&QA (Question Generation and Question Answering) generates several questions and answers related to the generated image using the textual input and employs VQA methods to compare and evaluate the image, such as T2I-CompBench \cite{huang2023t2icompbench}. While such methods have achieved some results in measuring the consistency between text and image content, they cannot effectively assess the model's overall performance under complex prompt conditions. 

We can derive two insights from the current evaluation methods. First, current methods focus on two different modalities, text input and image output, which inherently involve differences and information asymmetry between modalities, making evaluation challenging. Second, a good generative model should perform well in complex, real-world scenarios. Current methods focus on evaluating single attributes, but they lack comprehensive assessment for complex real-world scenarios involving multiple conditions.

Motivated by these insights, we propose \textbf{image regeneration} task, similar to "Painting Reproduction", given a reference image, we require the model under evaluation to generate an image based on it. The generated image is then assessed by comparing it to the reference image to determine the model's generation ability. \Cref{fig:comparch} compares the architecture of image regeneration and current metrics. The reference image is more informative than text prompts and aligns the output modalities, leading to more reasonable evaluation results. Since the generative model uses text inputs, we have developed an MLLM\cite{yang2023dawn,Nyberg_2021} based method to convert from image to text input. We propose a framework ImageRepainter for evaluating the quality of text-to-image models based on the task of image regeneration. The framework involves two stages. Specifically: \textbf{(1) Image understanding:} Firstly, based on MLLMs, the image information is organized to generate a tree-like structure called the image understanding tree (IUT), and then text prompts are generated using the information from it. \textbf{(2) Iterative generation:} Iterative exploration \cite{yang2023idea2img,yang2024mastering,wang2023diffusiondb} is typically involved in T2I generation for better images. This stage includes 4 parts: prompt generation/revision, image generation, image selection, and feedback generation. Furthermore, We introduce two benchmarks respectively designed for the evaluation of the content and style of the generated results.

The contributions if this work can be summarized as: 
\begin{itemize}
    \item \textbf{Novel T2I Model Evaluation Task: }This framework’s conception is rooted in the concept of “Painting Reproduction” which naturally aligns with the human judgment methodology. The task is able to measure the overall \textbf{generative capability} of the T2I model as well as reflect its \textbf{generative speciality}.
    \item \textbf{Effective Image Comprehensive Mechanism:} We introduce Image Understanding Tree(IUT) to enhance MLLMs' mutimodel ability. By defining fundamental rules, we incentivize MLLMs to interact with images and summarize the image information in a hierarchical tree structure, namely IUT. 
    \item \textbf{Two Diverse Benchmarks: }We propose two benchmarks respectively designed for the evaluation of the \textbf{content} and \textbf{style} of the generated results. The experimental results indicate that our benchmark aligns with human perception in assessing the generation capabilities of T2I models.
\end{itemize}

\section{Related Work}
\subsection{Evaluation of Generation Tasks}
Existing metrics for text-to-image generation can be categorized into fidelity assessment, alignment assessment, and LLM-based metrics \cite{huang2023t2icompbench}. Traditional metrics such as Inception Score (IS) \cite{salimans2016improved} and Frechet Inception Distance (FID) \cite{heusel2018gans} are commonly used to evaluate the fidelity of synthesized images. For evaluating image-text alignment, CLIP \cite{radford2021learning} and BLIP2 \cite{li2023blip2} are typically used for semantic matching between text and image. Recently, more and more work leverages the powerful reasoning capabilities of LLMs 
 \cite{yang2023idea2img,yang2024mastering} for evaluation. However, the measurement of text-to-image generation does not intuitively reflect human comparisons and perceptions of visual information. Therefore, we propose an evaluation framework for the text-to-image model based on the image regeneration task, simulating the form of human painting reproduction, which provides a natural and reasonable judgment for T2I model quality.

\subsection{LLM-driven Automatic System}
In the field of natural language processing (NLP), there has been a significant transformation with the emergence of LLMs \cite{chowdhery2022palm,ouyang2022training,touvron2023llama}, which have demonstrated significant capabilities in interacting with humans through conversational interfaces. The "Chain-of-Thought" (CoT) framework \cite{kojima2023large,wei2023chainofthought,zhang2022automatic} opens the door to further enhance the capabilities of LLMs, which guides LLMs to progressively generate answers, aiming to obtain higher-quality response. Recent research has pioneered novel methodologies by integrating external tools or models with Large Language Models (LLMs). For example, Toolformer \cite{schick2023toolformer} facilitates LLMs’ access to external tools via API tags. Visual ChatGPT \cite{wu2023visual} and HuggingGPT \cite{shen2023hugginggpt} have broadened the scope of LLMs by enabling them to leverage other models for tasks extending beyond linguistic domains. Furthermore, PromptBreeder \cite{fernando2023promptbreeder} and Idea2Img \cite{yang2023idea2img} frameworks respectively utilize LLMs for automated optimization of text prompt and image design. Inspired by these efforts, we embrace the concept of LLMs as multifunctional tools and utilize this paradigm to construct an iterative framework for evaluating the quality of generative models with the image regeneration task.

\section{Methodology}
\begin{figure*}[h]
  \centering
  \includegraphics[width=\linewidth]{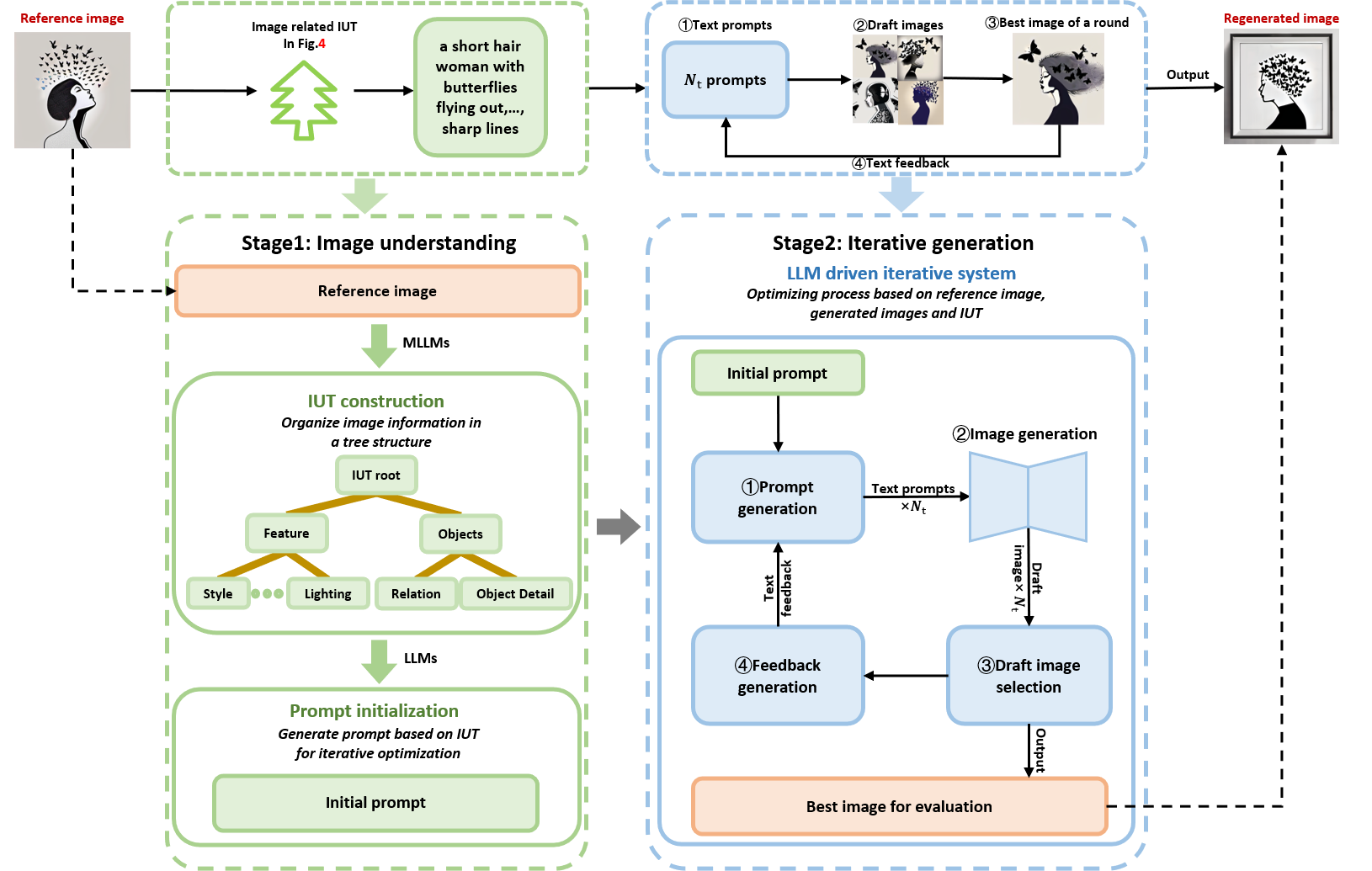}
  \caption{\textbf{Overview of ImageRepainter}. The framework consists of two stages: image understanding and iterative generation. These stages are displayed from left to right and interact continuously with LLM. The detailed process of each stage is shown above: \textbf{(1) Image understanding:} Firstly, the image information is organized to generate a tree-like structure called the image understanding tree (IUT), and then initial prompts are generated using the information from IUT. \textbf{(2)Iterative generation:} The second stage comprise four parts: prompt generation, image generation, image selection, and feedback generation.}
  \label{fig:framework}
\end{figure*}

We propose image regeneration task for T2I models evaluation. Similar to the concept of human "Painting Reproduction", it is easier for humans to make comparison and judgement within the same modality. We evaluate the generated images based on reference images to reflect the capability of the generative model. We introduces ImageRepainter, shown in \Cref{fig:framework}, a T2I model evaluation framework based on the image regeneration task. It employs MLLMs facilitating T2I models to generate images based on a specific reference image in terms of content, style, and other aspects. Finally, the framework utilizes the image-to-image metrics between the generated images and the reference images as the standard for evaluating the quality of the generation model.

\subsection{Image Understanding}
\begin{figure}[h]
  \centering
  \includegraphics[width=0.9\linewidth]{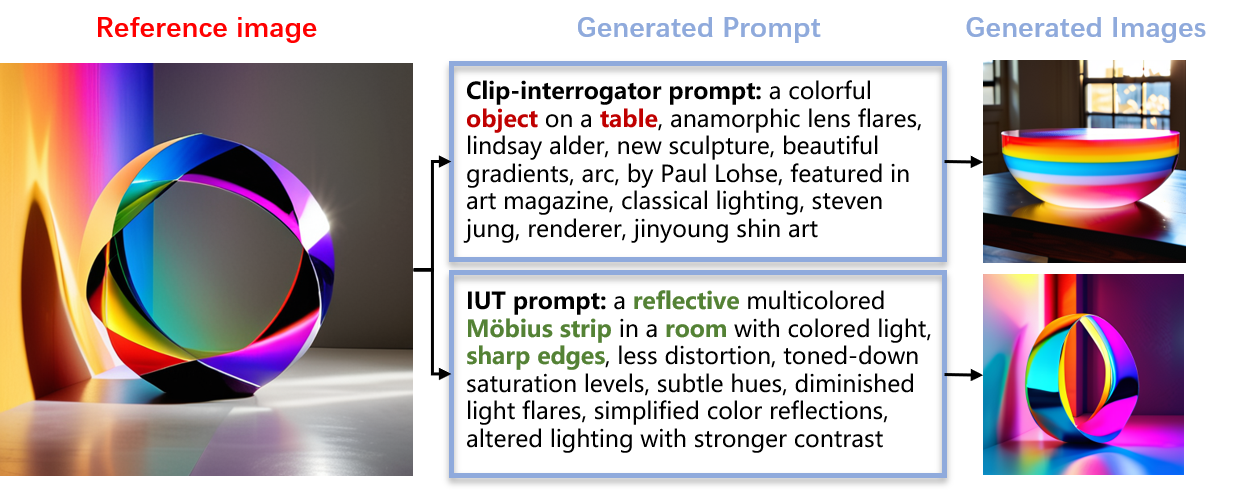}
  \caption{Examples of the generated images by using the prompt from CLIP-interrogator and our proposed IUT. We can observe that the accuracy of the information described in the prompts generated by the CLIP-interrogator is insufficient, leading to unsatisfactory results due to incomplete information.}
  \label{fig:CI}
  \vspace{-0.4cm}
\end{figure}

Image understanding is the first stage of this framework aiming at generating a high quality description of the reference image. The CLIP-interrogator model \cite{li2022blip,radford2021learning} in can generate stable diffusion prompts associated with the image input, thus producing image understanding. We directly employ the CLIP-interrogator for image regeneration tasks, as shown in \Cref{fig:CI}. The prompt produced may contain jumbled text and lacks accuracy. Therefore, the CLIP-interrogator cannot serve as a ideal method for image understanding. 

In this regard, we utilize MLLMs for image understanding, since they exhibit strong capabilities and align well with human cognition. We introduce Image Understanding Tree (IUT), in order to organize the information of an image in a tree structure, as this prevents redundancy and allows for a clear delineation of features at various levels of granularity. Constructing the IUT requires the use of multimodal large language model $M$(GPT4v) to analyze the reference image. We design templates to generate JSON format output for its standardization.
\begin{figure*}[h]
  \centering
  \includegraphics[width=0.95\linewidth]{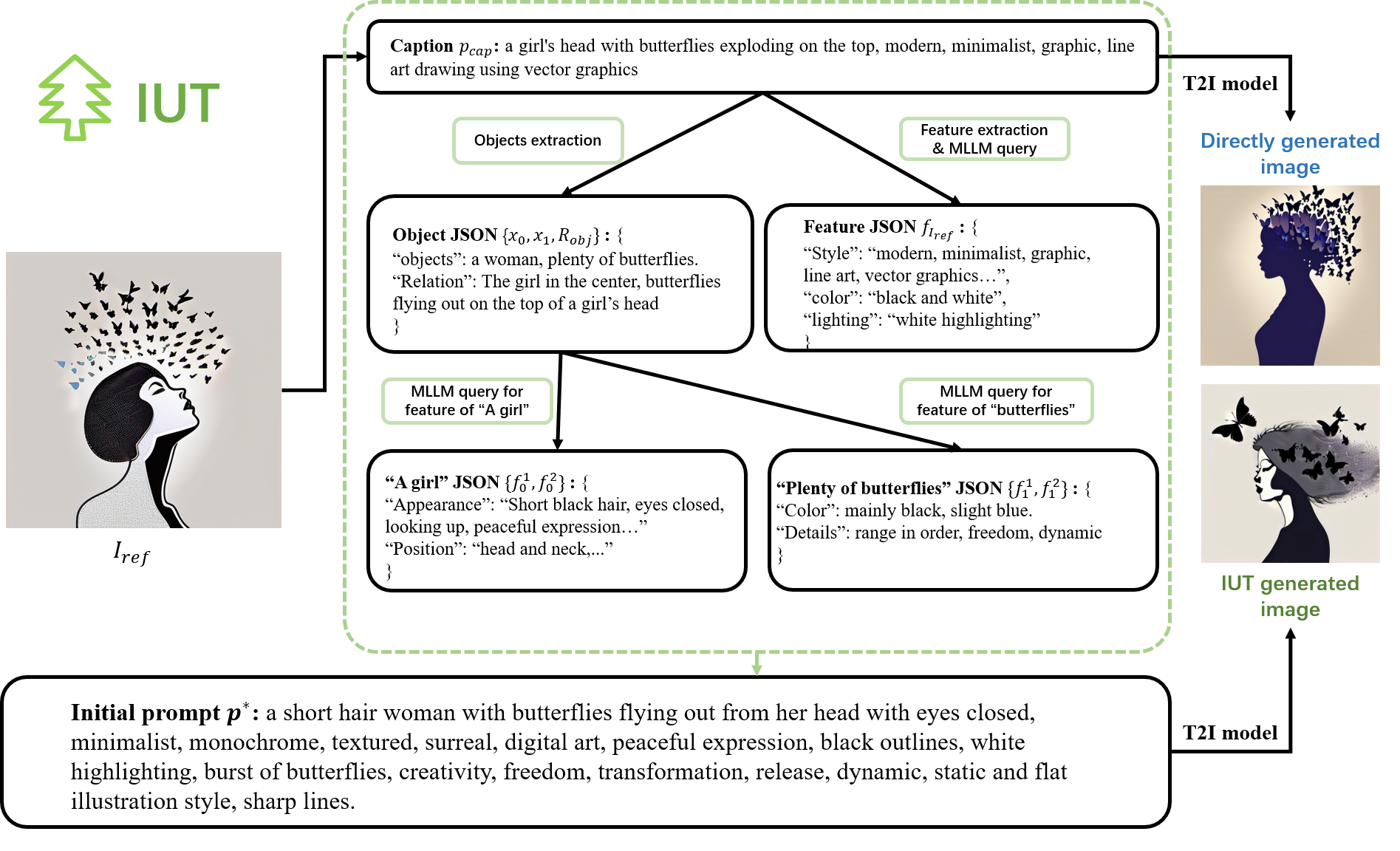}
  \caption{An example IUT construction, which shows that IUT capture more information such as color and facial details of the image than the direct caption.}
  \label{fig:IUT}
\vspace{-0.4cm}
\end{figure*}

As shown in \Cref{fig:IUT}, given a portrait image $I_{ref}$, the first step is to generate a caption $p_{cap}$ of the image as a base prompt. Then, we guide the MLLM to extract the overall features of the image $f_{I_{ref}}$, the objects within the image $\{x_0,...,x_{k-1}\}$, and the relationships between the objects $R_{obj}$ from the image using text template $T_{ext}$
\begin{equation}\label{eq2}
  \{f_{I_{ref}}, x_0,...,x_{k-1}, R_{obj}\} = M(I_{ref}, T_{ext}).
\end{equation}
Subsequently, more detailed information is extracted for each object in the image. After using LLM for automated questioning with text template $T_{obj}$, detailed information about the respective objects is obtained. 
\begin{equation}\label{eq1}
  \{f^0_{x_i},...,f^{n-1}_{x_i}\} = M(I_{ref}, x_i, T_{obj})
\end{equation}
, where $f^j_{x_i}$ represents the $j$th feature of the object $x_i$. Finally, the initial prompt $p^*$ is generated based on the IUT of the reference image using LLM.

\subsection{Iterative Generation}
After generating prompt $p^*$ from the image understanding module, for the initial iterations, the prompt generation module is responsible for expanding the prompt $p^*$ into $N_1$ synonymous prompts to participate in the iterations, where $N_t$ represents the number of the prompts in the $t^{th}$ iteration. The reason for generating multiple synonymous prompts is due to the T2I model's bias in understanding synonymous words, which can lead to situations where a good prompt generates a bad image.

The iterative process comprises 4 parts: image generation, image selection, feedback generation, and prompt revision. 

\noindent\textbf{Image generation.} In the $t^{th}$ iteration, the framework input the $N_t$ prompts obtained from the prompt generation part into the T2I model under evaluation, resulting in $N_t$ corresponding image outputs.

\noindent\textbf{Image selection}. CLIP, DINOv2, GPT4v is used to assess the similarity between $N_t$ images and the reference image, selecting the highest scoring image and prompt. If the current iteration t reaches the maximum iteration T, the images from this iteration are considered as the model’s regeneration result of the reference image; otherwise, the iteration continues. This part is designed to retain relatively stable performance and generate high-quality images. CLIP and DINOv2 metrics are capable of measuring coarse-grained semantic and visual information in images, while GPT4v captures fine-grained content and perceputual information.

\noindent\textbf{Feedback generation.} Feedback generation is to provide guidance for modifying the prompt. Text feedback $F$ is generated based on the differences between the current image and the reference image, as well as the previously constructed IUT. 

\noindent\textbf{Prompt revision.} For intermediate iteration $t$, prompt modification is carried out based on the best-performing prompt after evaluation in the current iteration and the text feedback $F$, generating $N_{t+1}$ prompts for image generation. In each iteration, we only modify one aspect of the prompt, since we observe in practice that it can make the LLM more “focused” and generate better prompt results. 

\begin{table*}
\small
\setlength{\abovecaptionskip}{0.cm}

  \begin{tabular}{lcccccccc}
  \toprule
 \multirow{2}*{\textbf{Model}}& \multicolumn{4}{c}{\textbf{Image Regeneration}} & \multicolumn{2}{c}{\textbf{User study}}&\multirow{2}*{\textbf{T2I-comp}}&\multirow{2}*{\textbf{HPSv2/Pickscore}}\\
    \cmidrule(r){2-5} \cmidrule(r){6-7}
    &\textbf{CLIP(\%)}& \textbf{DINO(\%)}&\textbf{GPT4-con}&\textbf{GPT4-per}& \textbf{consistency}& \textbf{perceptual}&\\
    \midrule
    SD1.4& 89.98& 88.37&0.5500 & 0.4460& 0.4216& 0.3682
&0.3080&0.2563/0.0871\\
    SD1.5& 90.17& 90.33&0.5660 & 0.4760& 0.4338& 0.3976
&0.3315&0.2584/0.1015\\
    SD1.5-DPO& 91.08& 93.57&0.5560 & 0.6840& 0.4544& 0.6388
&0.3392&0.2614/0.1390\\
    SD2.0& 90.79& 91.68&0.6060 & 0.5860& 0.4960& 0.4626
&0.3386&0.2577/0.0787\\
    SD2.0-Inpaint& 90.67& 92.32&0.6220 & 0.6020& 0.4892& 0.4920
&0.3560&0.2611/0.0854\\
    SDXL1.0& 90.17& 92.41&0.7600 & 0.6600& 0.6726& 0.6448
&\textbf{0.4091}&0.2565/0.0999\\
    Juggernautv1& 93.37& 95.27&0.7120 & 0.7740& 0.6472& 0.8072
&0.3476&0.2705/\textbf{0.2118}\\
    Juggernautv9& \textbf{93.79}& \textbf{95.34}&\textbf{0.7700} & \textbf{0.8820}& \textbf{0.7056}& \textbf{0.8590}&0.3764&\textbf{0.2731}/0.1967\\
    \bottomrule
  \end{tabular}
   \caption{The evaluation result of our proposed method image regeneration, T2I-CompBench, and user study. GPT-con and GPT-per represents the content consistency and perceptual quality evaluated by GPT4v.}
   \label{tab:main-result}
\end{table*}

\section{Experiments}
\subsection{Experiments Setting}
\noindent\textbf{Task Definition: }We use the Image Regeneration task to  evaluate T2I models. In the Image Regeneration task, the T2I model is required to generate an image according to a reference image, akin to the process of human image repainting. It provides a more intuitive assessment compared to directly evaluating the alignment between the T2I model’s input text and output image.

\noindent\textbf{Human evaluation: }When evaluating T2I models, the quality of the judgment method is determined by its alignment with human assessments. Our user study references the Likert scale human evaluation template ImagenHub \cite{ku2024imagenhubstandardizingevaluationconditional}. Specifically, we ask annotators to rate both the content consistency between the generated image and the text prompt, and the perceptual quality of the image, on a scale from 1 to 5. For each model, we randomly select 50 text-image pairs for evaluation, with each pair being rated by 5 human annotators. A total of 40 participants are involved, with each annotator rating 50 text-image pairs. We normalize and present the two-dimensional scores for each model in \Cref{tab:main-result}. 

\noindent\textbf{Baseline for T2I model evaluation: }We apply T2I-CompBench \cite{huang2023t2icompbench} 3-in-1 evaluation as the baseline for content consistency evaluation which takes attribute binding, object relationship into consideration. HPSv2 score and PickScore serve as the baseline for perceptual quality evaluation.

\noindent\textbf{Baseline for image understanding: }We apply CLIP-interrogator as the baseline for our proposed image understanding method. The CLIP-interrogator can generate text prompts for a given image input, which in turn can generate images similar to the input. In this experiment, the CLIP-interrogator is used as a baseline to assess whether the framework is capable of completing the image regeneration task. By quantitatively comparing with the evaluation metrics of the CLIP-interrogator, we aim to demonstrate that our proposed framework can provide a more accurate and superior understanding of images.

\noindent\textbf{Implementation Details: }We set iteration rounds $T=4$. We define a queue of iterative elements, where each round iterates over one element in the queue, including overall image, style, color, and detailed content in sequence.

\noindent\textbf{Models: }In the experiment, we utilize GPT4v \cite{openai2024gpt4} as MLLM and employ ChatGPT to handle pure text tasks in order to save resources, specifically the text-davinci-003 version. To control the LLM’s response, we utilize JSON format \cite{shen2023hugginggpt} to constrain the text output of the LLM. For the generation model used in our experiment, we employed the stable diffusion officially released model \textit{SD1.4}, \textit{ SD1.5}, \textit{SD2.0}, \textit{SD2.0-inpainting} \cite{Rombach_2022_CVPR}, the state-of-the-art \textit{SDXL1.0} \cite{podell2023sdxl}, and popular models from the open community to assess the quality of the generation model within our framework. In comparison to the officially released base models, custom models can generate relatively stable quality images. However, due to fine-tuning, the diversity in content or style generated by these models might be relatively diminished. Using multiple models can demonstrate the effectiveness of our proposed evaluation framework. \textit{SD1.5-DPO} \cite{Wallace2023DiffusionMA} is a fine-tuned version of \textit{SD1.5 }by directly optimizing on human comparison data. 
The $JuggernautXL_{v9}$ and $JuggernautXL_{v1}$ \cite{juggernautxl} are the most popular and effective models from the Civitai community, with the former being the latest version and the latter being the original version. 

\begin{figure}[h]
  \centering
  \setlength{\abovecaptionskip}{0.2cm}
  \includegraphics[width=0.85\linewidth]{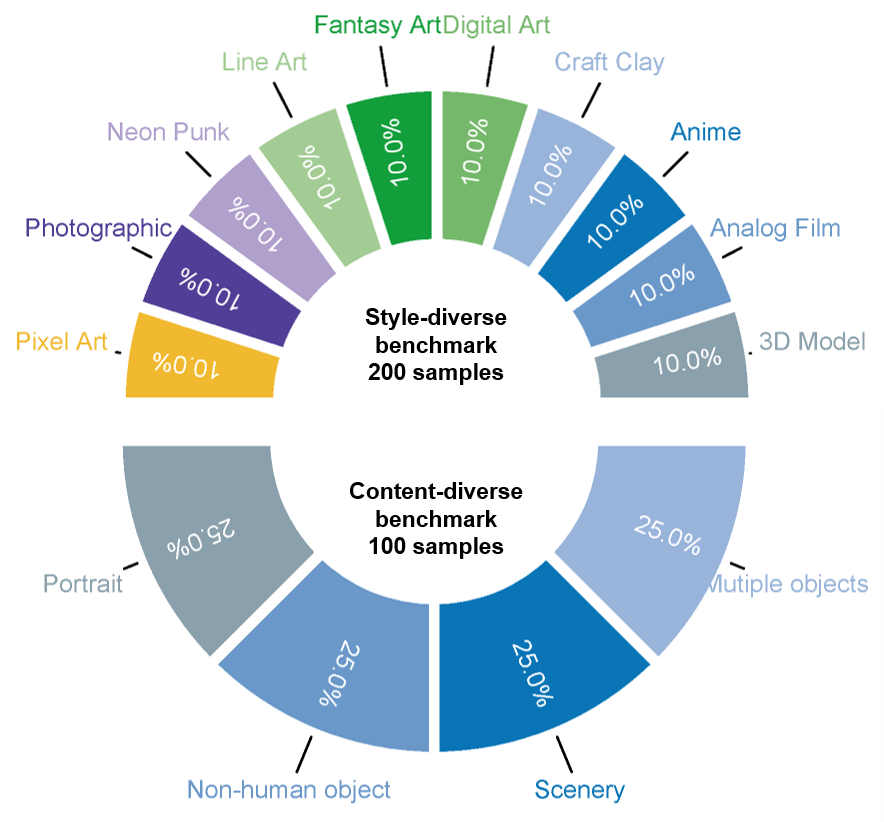}
  \caption{The distribution of style-diverse benchmark and content-diverse benchmark. }
  \label{fig:distribution}
\end{figure}

\noindent\textbf{Evaluation Datasets: }For quantitative evaluation, we constructed two benchmarks, respectively designed for the evaluation of content and style of the generated results. The \textbf{style-diverse benchmark} consists of 200 text-image samples with 10 different styles. The \textbf{content-diverse benchmark} consists of 100 samples with 4 different types of content. The style-diverse benchmark is composed of 20 manually selected captions combined with descriptions of 10 style categories, and normalized and synonym-transformed using ChatGPT. The content-diverse benchmark, on the other hand, is manually collected through an open-source creation platform and normalized using ChatGPT. The data distribution of the two benchmarks is illustrated in \Cref{fig:distribution}.

\noindent\textbf{Evaluation Metrics: }We use CLIP \cite{radford2021learning}, DINOv2 \cite{oquab2024dinov2}, and GPT4v. CLIP score is capable of evaluating the semantic information between images. DINOv2 metric, compared to the CLIP metric, is more sensitive to visual information such as lighting and color tones in images. 
GPT-4V scores are given on a scale from 1 to 5 for two dimensions: content consistency between the reference image and the generated image, and perceptual quality.

\subsection{Evaluating T2I Models}
We use the content-diverse benchmark for evaluation. The results are shown in \Cref{tab:main-result}. 
It can be observed that using image regeneration for evaluation aligns more closely with human annotations. In terms of content consistency, the SDXL1.0 model performs best in the T2I-Compbench evaluation, though it differs from human judgments. On the perceptual level, Image Regeneration method outperforms PickScore. Both our method and HPSv2 scores align with human judgments, but our method better reflects the perceptual quality differences of the models and also provides interpretable text (provided in supplementary materials).

\begin{figure}[h]
  \centering
  \setlength{\abovecaptionskip}{0.2cm}
  \includegraphics[width=\linewidth]{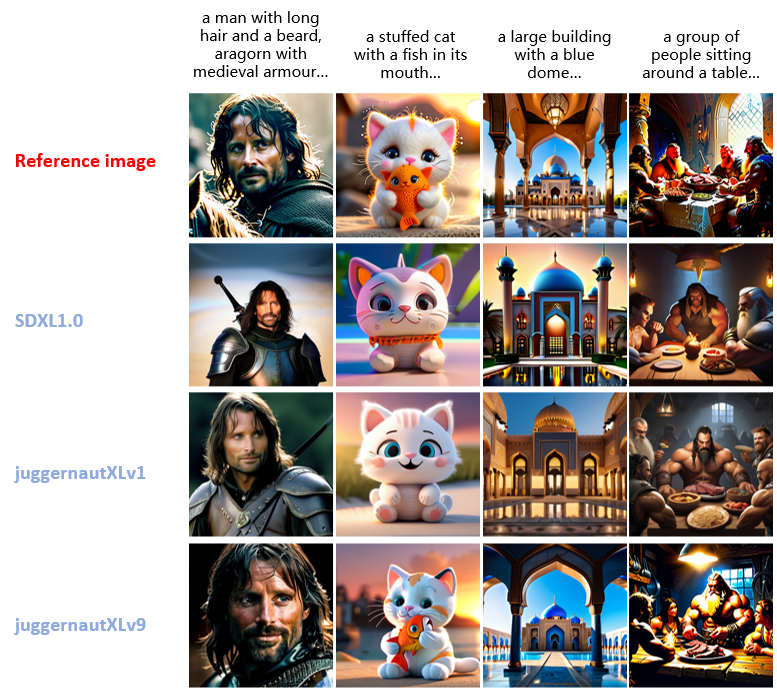}
  \caption{Cases intuitively demonstrate the generation capabilities of T2I models, showing that our evaluation of the four T2I models is reasonable. }
  \label{fig:main_comp}
\end{figure}

To more intuitively demonstrate the model's comprehensive generation capabilities, we list random selected cases of Image Regeneration method. As shown in \Cref{fig:main_comp}, based on visual perception, it can be observed that $JuggernautXL_{v9}$ exhibits the strongest generalization and generation capabilities, consistent with our proposed Image Regeneration assessment results, which indicates that our proposed evaluating method better aligns with human cognition.

\begin{figure}[h]
\centering
\subfigure[CLIP: Style evaluation]{
\begin{minipage}[t]{0.48\linewidth}
\centering
\includegraphics[width=\linewidth]{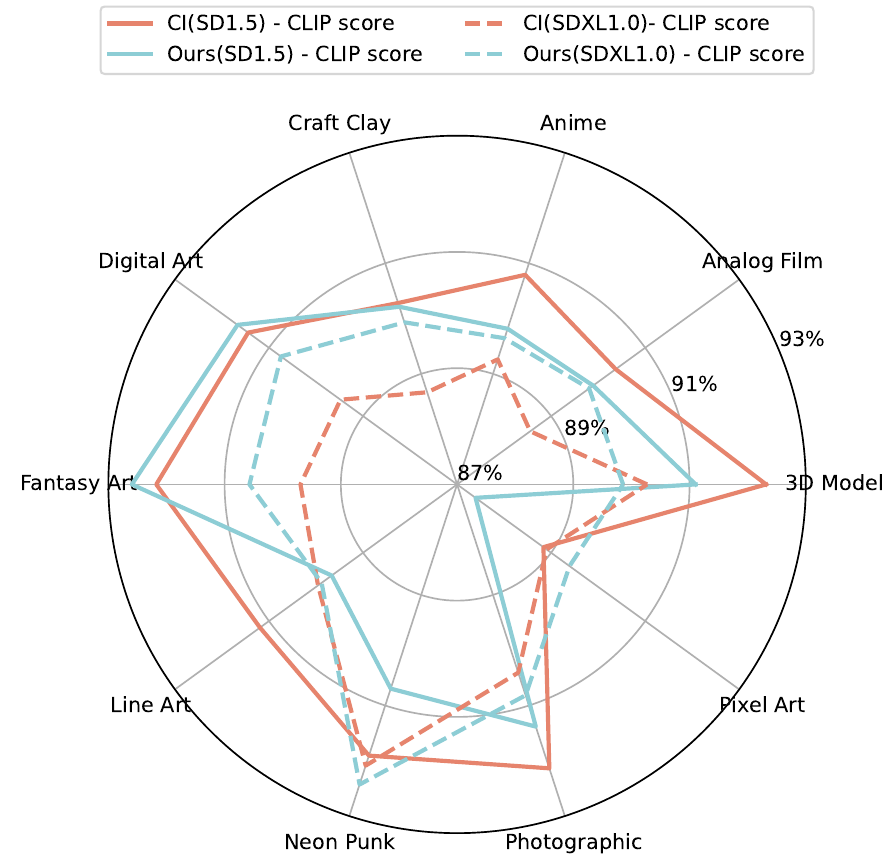}
\end{minipage}%
}%
\subfigure[DINO: Style evaluation]{
\begin{minipage}[t]{0.48\linewidth}
\centering
\includegraphics[width=\linewidth]{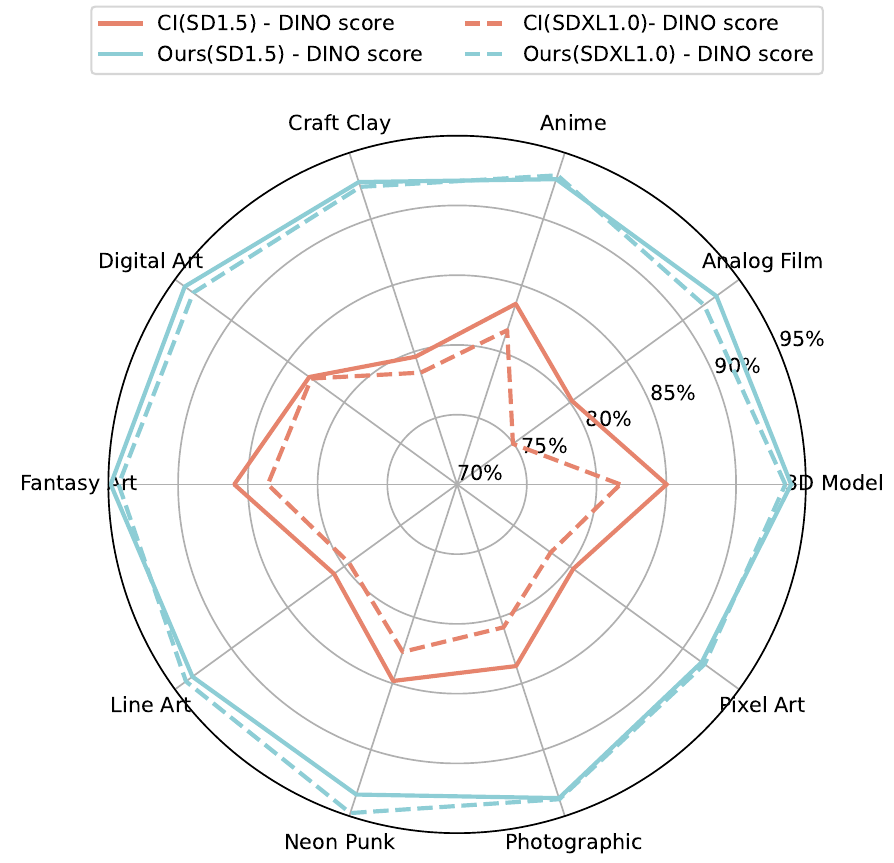}
\end{minipage}%
}%

\subfigure[GPT4v: Style evaluation]{
\begin{minipage}[t]{0.48\linewidth}
\centering
\includegraphics[width=\linewidth]{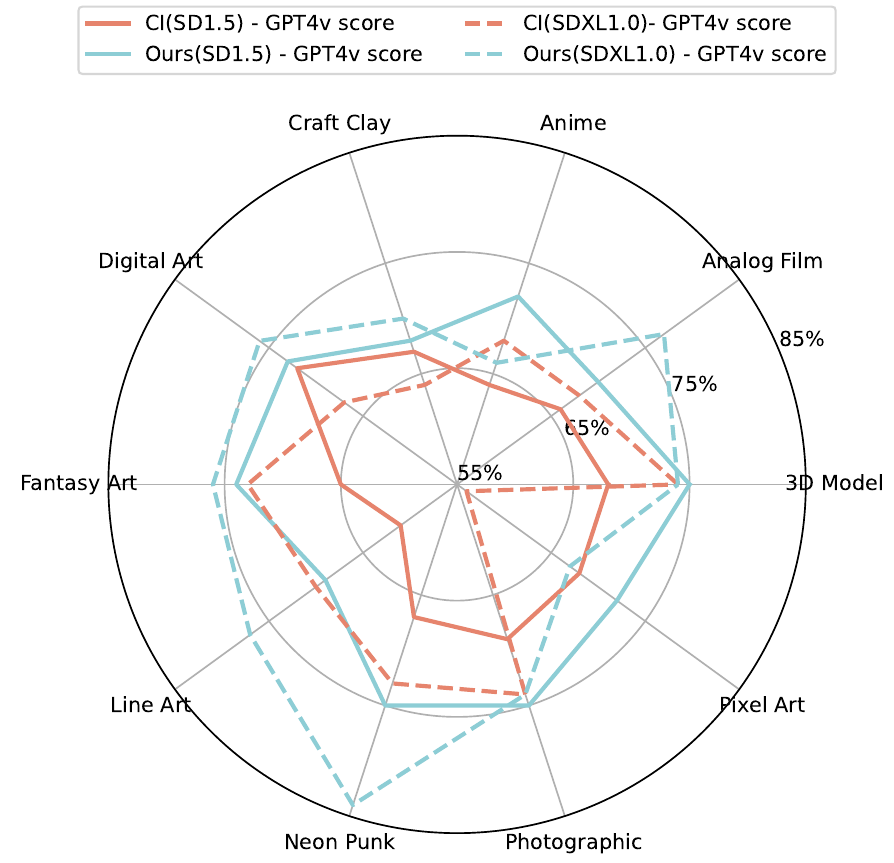}
\end{minipage}
}%
\subfigure[CLIP: Content evaluation]{
\begin{minipage}[t]{0.48\linewidth}
\centering
\includegraphics[width=\linewidth]{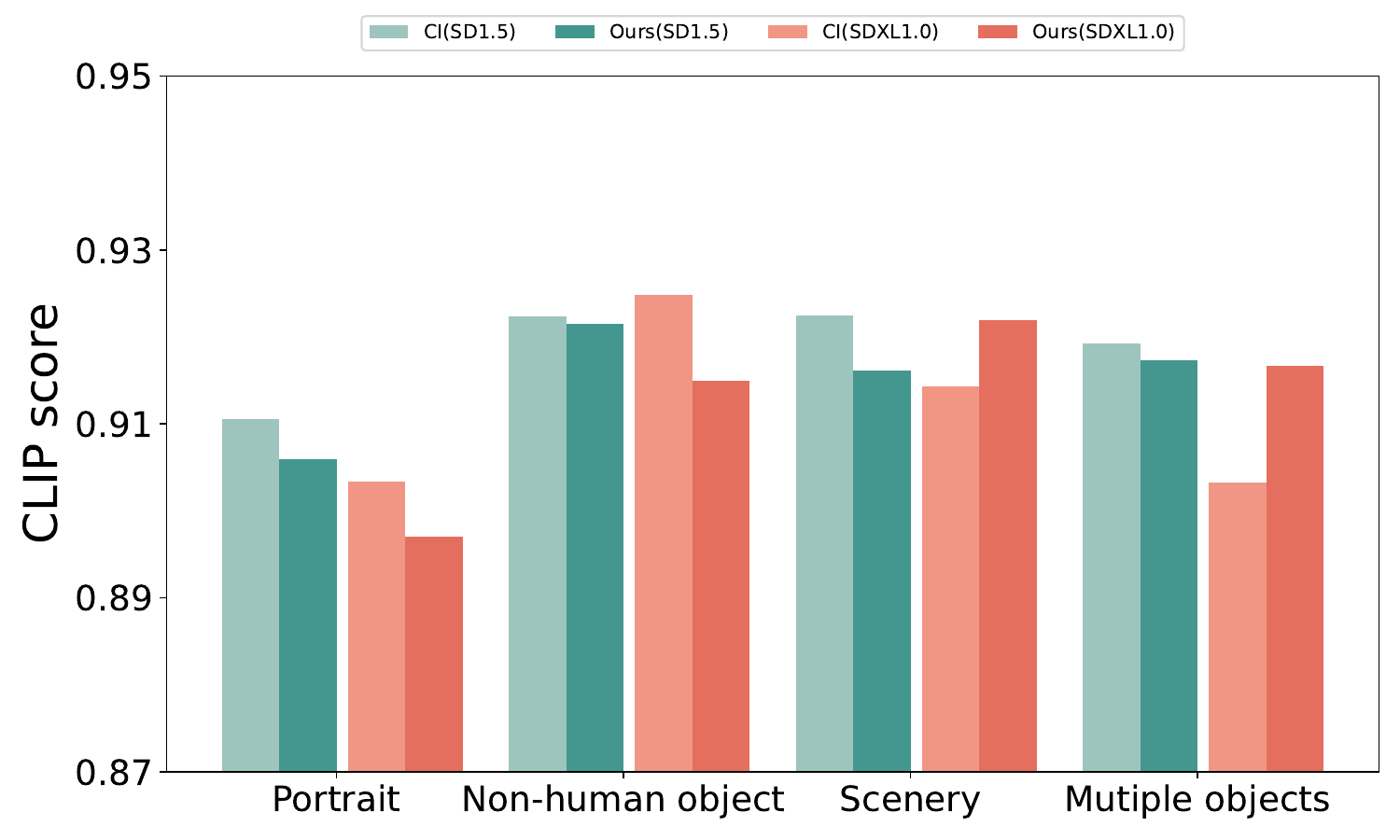}
\end{minipage}%
}%

\subfigure[DINO: Content evaluation]{
\begin{minipage}[t]{0.48\linewidth}
\centering
\includegraphics[width=\linewidth]{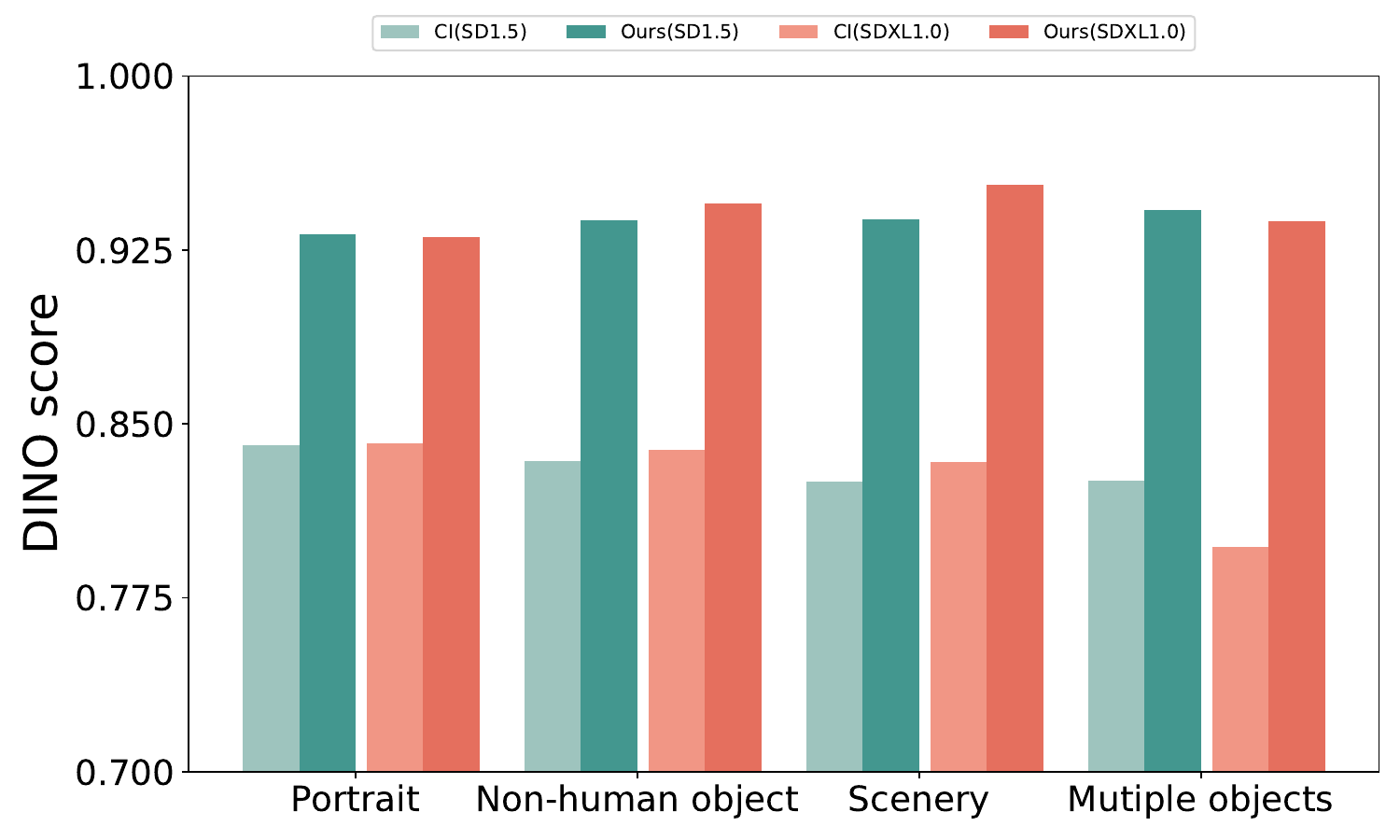}
\end{minipage}%
}%
\subfigure[GPT4v: Content evaluation]{
\begin{minipage}[t]{0.48\linewidth}
\centering
\includegraphics[width=\linewidth]{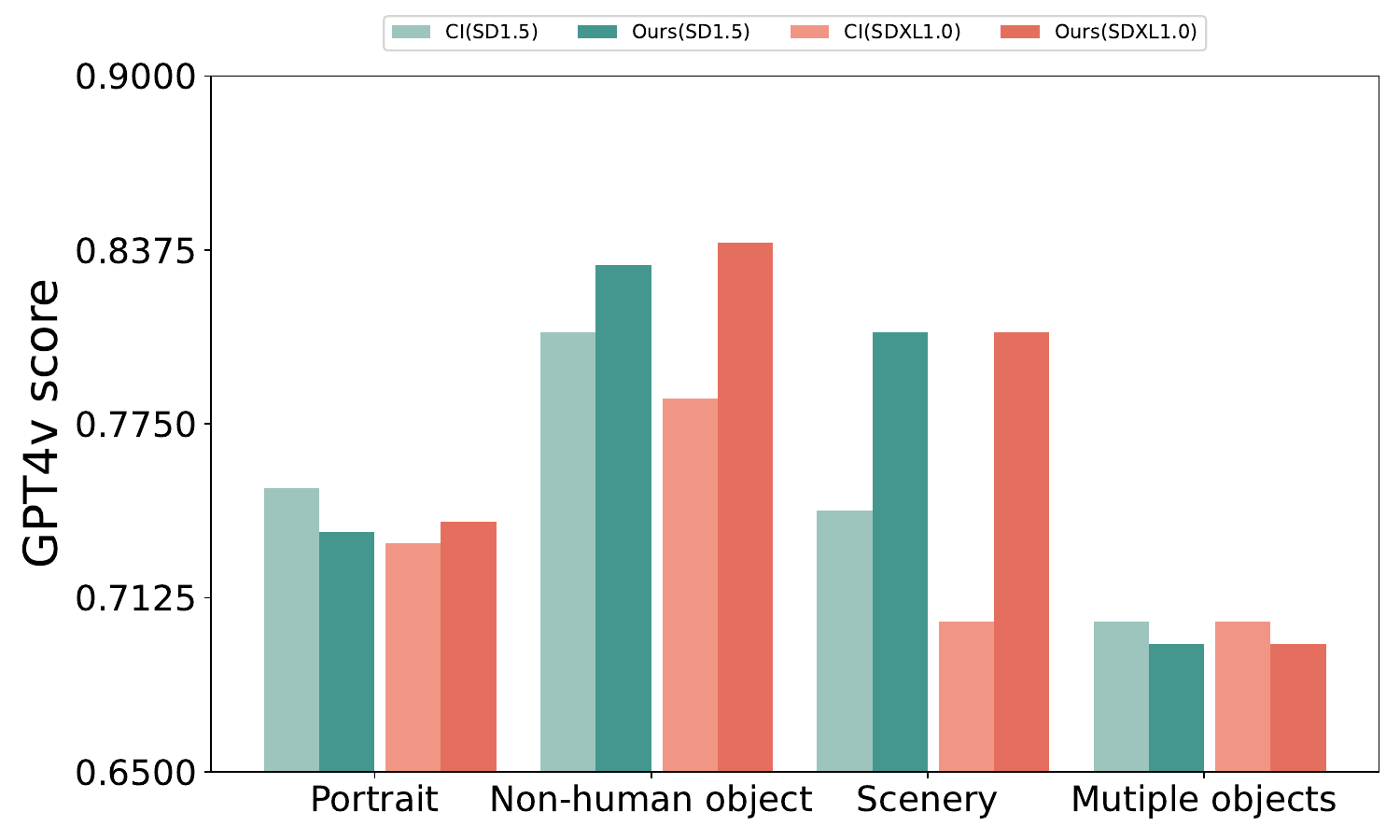}
\end{minipage}
}%
\centering
\caption{ The ImageRepainter performs better than the CLIP-interrogator across various styles. There is a significant improvement in both the DINOv2 and GPTv4 metrics.}
\label{fig:CI-OUR comparison}
\vspace{-0.4cm}
\end{figure}

\subsection{Enhancement of Image Understanding}
In order to prove that ImageRepainter has a correct and superior understanding of images, we performed image regeneration experiments using CLIP-interrogator as a control. We utilize content-diverse and style-diverse datasets to evaluate ImageRepainter’s performance across various types of image inputs.
We employ \textit{SD1.5} and \textit{SDXL1.0} as the T2I model in the framework. \Cref{fig:CI-OUR comparison} compares the CLIP, DINOv2, GPT4v metrics of the images generated by the CLIP-interrogator and ImageRepainter.

We can observe that using our framework for image regeneration tasks, the images obtained have a significant advantage over those obtained through the CLIP-interrogator for caption-based images, as judged by the criteria of DINOv2 and GPT4v. This demonstrates the effectiveness and superiority of employing the LLM-driven image regeneration and prompt iteration. 

Furthermore, we can observe models' generation speciality from the results, which indicate that the \textit{SD1.5 }model and the \textit{SDXL1.0} model perform relatively evenly across various type of styles. As for content-diverse evaluation, we observe that both the \textit{SD1.5} and \textit{SDXL1.0 }models have limited abilities in generating portraits and multi-object object images. Despite our framework's ability to iteratively generate better descriptions for images, the generation results remain unsatisfactory. 

\subsection{Ablation Study}
We conduct ablation study on Image Regeneration task itself, IUT, and iterative process.

\begin{figure}[h]
  \centering
  \includegraphics[width=\linewidth]{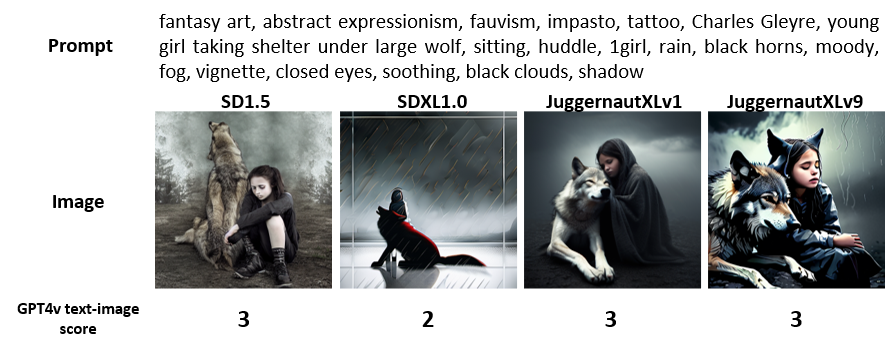}
  \caption{Direct text-image evalution via GPT4v.}
  \label{fig:textimg}
\end{figure}

\noindent\textbf{Effectiveness of Image Regeneration: }To prove the effectiveness of our method, we conduct experiments on directly using GPT-4v for text-image matching on content-diverse dataset. As shown in \textbf{\Cref{tab:textimg}} and \textbf{\Cref{fig:textimg}}, we can observe that directly using GPT-4v for text-image matching \textbf{lacks distinguishability} of different models compare to Image Regeneration. It may stem from the fact that MLLM performs better when comparing within the same modality, while it cannot fully leverage the capabilities of large models for different modalities such as image and text. It also demonstrates the potential of the Image Regeneration method alongside the development of MLLMs.
\begin{table}[h]
\small  \begin{tabular}{lccc}
    \toprule
    Model&Direct GPT4v&Image Regeneration&User study\\
    \midrule
    SD1.5& 0.6620& 0.5210&0.4157\\
 SDXL1.0& \textbf{0.7440}& 0.7100&0.6587\\
 juggerv1& 0.6960& 0.7430&0.7272\\
    juggerv9& 0.7220& \textbf{0.8260}&\textbf{0.7823}\\
    \bottomrule
\end{tabular}
\caption{Comparison between the Image Regeneration method and direct employment of GPT-4v for text-image evaluation.}
\label{tab:textimg}
\vspace{-0.4cm}
\end{table}

\noindent\textbf{Impact of IUT: }In this section, we designed experiments to verify the effectiveness of the proposed IUT on image understanding and generation. \Cref{tab:ablation} presents the results of prompt generation directly from images and prompt generation incorporating IUT. The experimental results indicate that IUT is more effective to relatively high quality models, since they exhibit strong and accurate execution of input text.
\begin{table}[h]
\scriptsize

  \begin{tabular}{l||@{\hspace{3pt}}l@{\hspace{9pt}}l@{\hspace{6pt}}l@{\hspace{6pt}}l}
    \hline 
    \textbf{Model}& \textbf{Method}& \textbf{CLIP(\%)}& \textbf{DINO(\%)}&\textbf{GPT4v(\%)}\\ 
    \hline
    \multirow{2}{*}{$JuggerXL_{v9}$}& Direct& 93.66& 94.83& 76.8\\ 
    & IUT& \textbf{95.71}\textcolor{deepgreen}{($+2.05$)}& \textbf{95.71}\textcolor{deepgreen}{($+0.88$)}& \textbf{84.4}\textcolor{deepgreen}{($+7.6$)}\\ 
    \hline
    \multirow{2}{*}{$SDXL1.0$}& Direct& \textbf{90.18}& 90.41& 62.8\\
    & IUT& 90.17\textcolor{red}{($-0.01$)}& \textbf{91.48}\textcolor{deepgreen}{($+1.07$)}& \textbf{66.0}\textcolor{deepgreen}{($+3.2$)}\\
  \hline
  \end{tabular}
\caption{Quantitative results on the effectiveness of our proposed image understanding template IUT. Method "Direct" generates the initial prompt directly from the reference image using MLLM. Method "IUT" generates the initial prompt based on IUT.}
 \label{tab:ablation}
 \vspace{-0.4cm}
\end{table}

\begin{figure}[h]
  \centering
  \includegraphics[width=\linewidth]{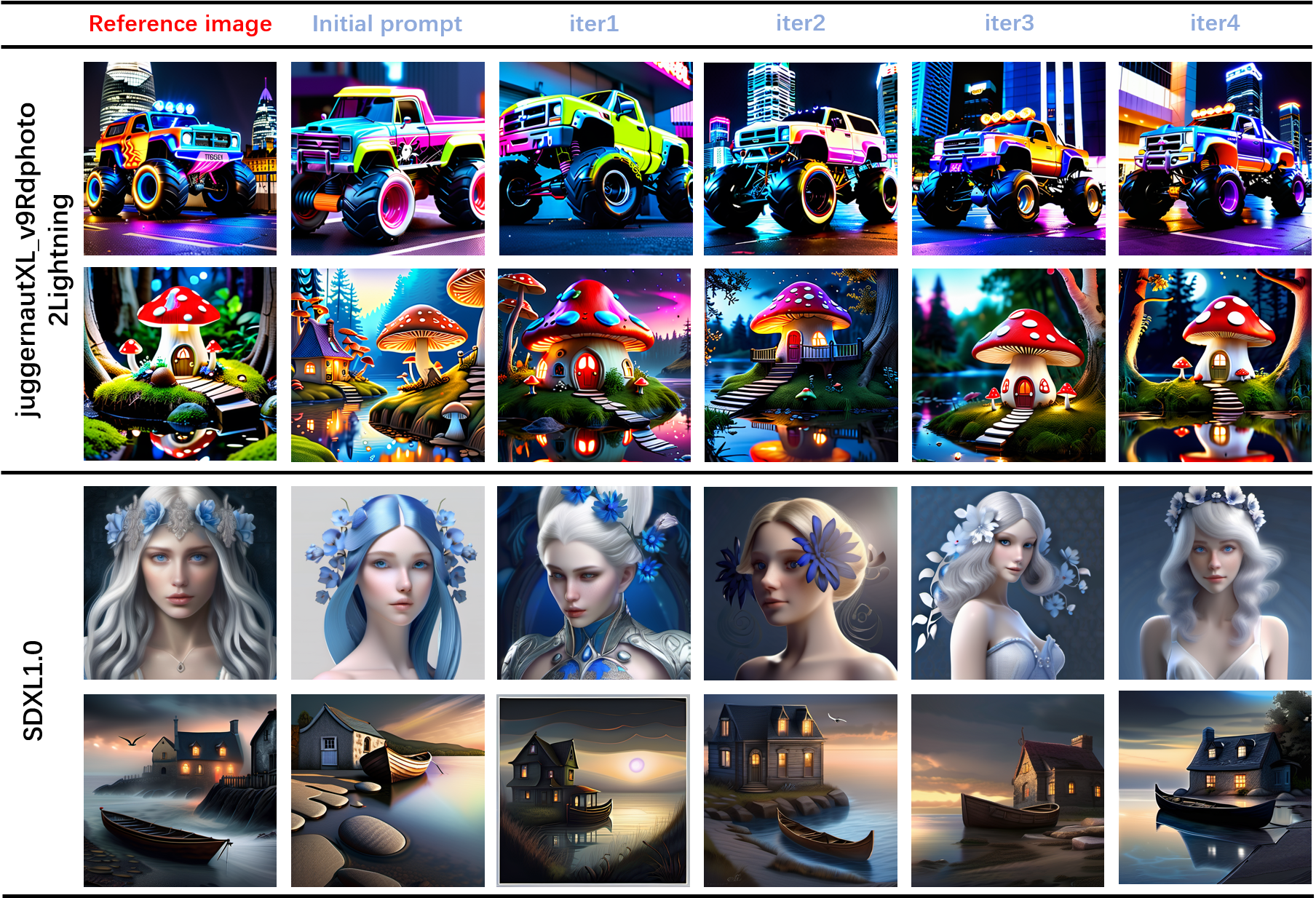}
  \caption{Ablation study of iteration rounds. The result indicates that the iteration plays a more important role in the models with relatively low quality. The models with high-quality can generate similar and good images with fewer iterations.}
  \label{fig:iter}

\vspace{-0.4cm}
\end{figure}

\noindent\textbf{Impact of Iteration Rounds: }In this section, we qualitatively demonstrate the impact of iteration rounds on the quality of image generation. \Cref{fig:iter} presents several examples. We can observe that when using the higher-quality $JuggernautXL_{v9}$ model, the iterations have a less noticeable impact on the improvement of image quality. Conversely, for the relatively weaker quality\textit{ SDXL1.0} model, iterations have a more significant impact on the image quality improvement. This is because the higher-quality model generates images more consistently, while \textit{SDXL1.0} may be influenced more by the seed and requires repeated iterations to obtain a relatively stable prompt performance.

\section{Conclusion}
In this paper, to fill the research gap in the evaluation of generative models, we propose ImageRepainter, an LLM-driven framework for assessing the quality of text-to-image models with\textbf{ image regeneration }task and introduce ImageRepainter framework. The framework iteratively generate high-quality images to explore the generation capability of T2I models and evaluate the model's generation effectiveness based on a visual-to-visual intuitive understanding. The visual-to-visual assessment in this paper is better compared to current text-to-visual assessments because the former is more insensitive to fine-grained information and relatively intuitive in terms of human perception. Additionally, our ImageRepainter can contribute to the AIGC community, facilitating creative work, and can also be applied to other meaningful tasks such as dataset augmentation, demonstrating strong extensibility. For future work, we will continue to improve this new framework to incorporate more complex input conditions and further explore effective methods for assessing the quality of generative models.

\bibliography{aaai25}

\end{document}